\documentclass[letterpaper, 10 pt, conference]{ieeeconf}  

\IEEEoverridecommandlockouts                              

\title{\LARGE \bf \texttt{STORM}: \texttt{S}lot-based \texttt{T}ask-aware \texttt{O}bject-centric \texttt{R}epresentation for robotic \texttt{M}anipulation}

\author{Alexandre Chapin$^{1}$, Emmanuel Dellandrea$^{1}$ and Liming Chen$^{1}$%
\thanks{$^{1}$Ecole Centrale de Lyon, LIRIS, 69130, Ecully, France
        {\tt\small alexandre.chapin@ec-lyon.fr}}%
}

\usepackage[pagebackref,breaklinks,colorlinks]{hyperref}
\usepackage[T1]{fontenc}
\usepackage{graphicx}
\usepackage{multirow}
\usepackage{amssymb}
\usepackage{amsmath}
\usepackage{float}
\usepackage[table]{xcolor}
\usepackage{booktabs}
\usepackage{subcaption}
\usepackage{pifont}

\colorlet{posbase}{green!60!black}
\colorlet{negbase}{red}

\newcommand{\gradcolor}[1]{%
    \ifdim #1pt > 0pt
        \color{posbase!\fpeval{min(100, #1 * 30)}!black}%
    \else
        \color{negbase!\fpeval{min(100, abs(#1) * 30)}!black}%
    \fi
}

\newcommand{\cmark}{\ding{51}} 
\newcommand{\xmark}{\ding{55}} 

\begin{document}

\maketitle
\thispagestyle{empty}
\pagestyle{empty}

\begin{abstract}
Visual foundation models provide strong perceptual features for robotics, but their dense representations lack explicit object-level structure, limiting robustness and controllability in manipulation tasks. We propose STORM (\textbf{S}lot-based \textbf{T}ask-aware \textbf{O}bject-centric \textbf{R}epresentation for robotic \textbf{M}anipulation), a lightweight object-centric adaptation module that augments frozen visual foundation models with a small set of task-aware slots for robotic manipulation. Rather than fully tuning large backbones on the task, STORM employs an efficient two-stage training strategy: few layers of object-centric representation are first trained  on top of the frozen backbone through visual–semantic pretraining using language embeddings, then jointly adapted with a downstream manipulation policy for task alignement. This staged learning prevents degenerate slot formation and preserves semantic consistency while aligning perception with task objectives. Experiments on object discovery benchmarks and robotic manipulation tasks show that STORM improves control performance and generalization to visual shifts (distractors, textures, lighting) compared to directly using frozen or fine-tuned foundation model features, or existing object-centric representations. STORM serves not only as an efficient mechanism for refining generic foundation model features, but also as a novel way of injecting beneficial structural and semantic bias into policy learning.
\end{abstract}


\section{Introduction}
\label{sec:intro}
Robotic manipulation requires perceptual representations that capture fine-grained spatial structure while remaining interpretable and controllable by high-level task specifications. Recent Visual Foundation Models (VFMs) such as DINOv2~\cite{oquab2024dinov2learningrobustvisual} provide powerful dense feature representations that generalize across domains, but they do not expose explicit object-level structure. As a result, downstream policies must implicitly learn to attend to relevant entities, often leading to brittle behavior and poor generalization under visual variations~\cite{burns2023makespretrainedvisualrepresentations,chapin2025objectcentricrepresentationsimprovepolicy}.

Slot-based object-centric representation offers an appealing alternative. By decomposing scenes into discrete latent "slots", each representing an individual object or part, these representations promote modular reasoning, interpretable structure and compositional generalization. Frameworks such as Slot-Attention~\cite{locatello2020objectcentriclearningslotattention} and its successors~\cite{Didolkar2024ZeroShotOCRL,seitzer2023bridginggaprealworldobjectcentric,singh2022illiteratedallelearnscompose} have shown that meaningful object representations can emerge without supervision when proper inductive biases are applied. However, these models remain \textbf{unguided}: the slot formation process is purely visual, without semantic control. As a result, the learned slots may not correspond to \textbf{task-relevant entities} and may even merge into uncorrelated components~\cite{chapin2025objectcentricrepresentationsimprovepolicy}, limiting their usefulness for robotic manipulation. Recent efforts have attempted to integrate semantics into slot-based learning.  Shatter \& Gather~\cite{kim2023shattergatherlearningreferring} achieved strong alignment between slots and text post hoc through contrastive learning. CTRL-O~\cite{didolkar2025ctrlolanguagecontrollableobjectcentricvisual} successfully demonstrated language controllability, but its training relies on large language models (LLM2Vec~\cite{behnamghader2024llm2veclargelanguagemodels} and CLIP~\cite{radford2021learningtransferablevisualmodels}). While effective for object retrieval and image-generation conditioning, these approaches have yet to be adapted for robotic control scenarios, where aligning visual perception with instruction semantics is essential for manipulation. 

Our work presents three main contributions:
\begin{itemize}
    \item \textbf{A lightweight, slot-based and semantic-aware architecture (STORM)}: We propose a compact, stable object-centric module consisting of a small number of slot-based layers placed on top of frozen VFM features with a novel mask entropy loss to avoid degenerate solutions. Despite minimal architectural overhead, STORM translates dense visual features into semantically grounded, interpretable object representations directly consumable by control policies.

    \item \textbf{A two-phase adaptation strategy for reliable control}: We introduce a two-stage learning strategy to adapt VFM frozen features into task-relevant representations. By first stabilizing slot formation through visual-semantic pretraining and subsequently jointly training with a control policy, this approach effectively injects beneficial structural and semantic bias. Crucially, this process makes the visual representations task-aware, successfully preventing degenerate slot formation during policy training.

    \item \textbf{Simulated and real-world validation}: The performance of STORM is demonstrated through extensive experimental validation in both simulated environments and real-world robotic deployments. Our results demonstrate that it outperforms directly using frozen VFM features, full fine-tuning, or end-to-end object-centric training, showing improvements in generalization to visual distractors and overall control reliability.
\end{itemize}

\section{Related works}
\label{sec:related}

\subsection{Pre-trained visual foundation models}
Recent advances in visual representation learning have been driven by large-scale, self-supervised methods trained on internet-scale, unlabeled data. Many of these methods adopt Vision Transformer (ViT) architectures~\cite{dosovitskiy2021imageworth16x16words} and training objectives such as masked image modeling~\cite{he2021maskedautoencodersscalablevision}, contrastive learning~\cite{caron2021unsupervisedlearningvisualfeatures,caron2021emergingpropertiesselfsupervisedvision,chen2021empiricalstudytrainingselfsupervised}, or hybrids that combine both paradigms~\cite{oquab2024dinov2learningrobustvisual,zhou2022ibotimagebertpretraining}. The outcome of this line of work are highly robust pre-trained vision foundation models (VFMs), for example DINOv2~\cite{oquab2024dinov2learningrobustvisual}, that produce rich spatially-structured features well suited for downstream perception and reasoning. At the same time, multimodal training on paired image–text corpora has produced aligned vision–language models such as CLIP~\cite{radford2021learningtransferablevisualmodels}, enabling open-vocabulary and semantic supervision for visual systems.

Most robotic pipelines today rely on VFMs for sample-efficient policy learning~\cite{qian2024taskorientedhierarchicalobjectdecomposition,qian2024recastinggenericpretrainedvision,shi2024composingpretrainedobjectcentricrepresentations}. However, directly using dense, high-dimensional VFM features for control can be computationally expensive and may include task-irrelevant information. In contrast, STORM produces compact, discrete slot-based latents that are both task-aware and immediately usable by downstream policies. 

\subsection{Object-centric representation for robotics}
Object-Centric Representation Learning (OCRL) decomposes scenes into modular latent slots to promote compositionality in perception. While early methods struggled with real-world imagery \cite{locatello2020objectcentriclearningslotattention}, recent advances leverage powerful visual backbones \cite{seitzer2023bridginggaprealworldobjectcentric,singh2025glassguidedlatentslot,singh2022illiteratedallelearnscompose,singh2022simpleunsupervisedobjectcentriclearning,jiang2023objectcentricslotdiffusion,wu2023slotdiffusionobjectcentricgenerativemodeling} and vision-language models \cite{fan2024unsupervisedopenvocabularyobjectlocalization,kim2023shattergatherlearningreferring,xu2022groupvitsemanticsegmentationemerges} to semantically ground these slots. However, these methods predominantly focus on static tasks, and ensuring stable, task-relevant slot assignments often requires complex, multi-model pipelines \cite{didolkar2025ctrlolanguagecontrollableobjectcentricvisual,behnamghader2024llm2veclargelanguagemodels}. In robotics, object abstractions are increasingly adopted to improve policy generalization \cite{brohan2023rt2visionlanguageactionmodelstransfer,chapin2025objectcentricrepresentationsimprovepolicy,wen2025datacentricrevisitpretrainedvision,zhu2023learninggeneralizablemanipulationpolicies}, frequently utilizing pre-trained mask models like SAM \cite{kirillov2023segment,qian2024taskorientedhierarchicalobjectdecomposition,shi2024composingpretrainedobjectcentricrepresentations}. Yet, mask-based approaches can be brittle in clutter and rely on heuristic post-processing. Conversely, continuous slot-based representations are directly consumable by policies but often fail under visual distraction without explicit slot formation control \cite{chapin2025objectcentricrepresentationsimprovepolicy}. To bridge the gap between static semantic discovery and dynamic closed-loop control, STORM explicitly conditions slot formation on task instructions. By introducing a two-phase strategy that first stabilizes semantic object discovery and subsequently aligns these representations with downstream control objectives, our approach ensures robust, task-relevant abstractions for robotic manipulation.

\section{Method}
\label{sec:method}

\begin{figure*}[htbp]
    \centering
    \begin{subfigure}[b]{0.38\textwidth}
        \centering
        \includegraphics[width=\textwidth]{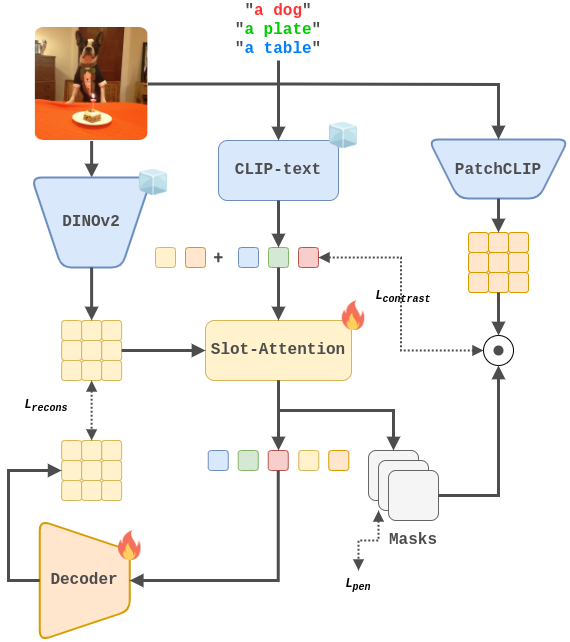}
        \caption{\textbf{Step 1: Semantic learning}}
        \label{fig:method_pt}
    \end{subfigure}
    \hfill
    \vrule
    \hfill
    \begin{subfigure}[b]{0.58\textwidth}
        \centering
        \includegraphics[width=\textwidth]{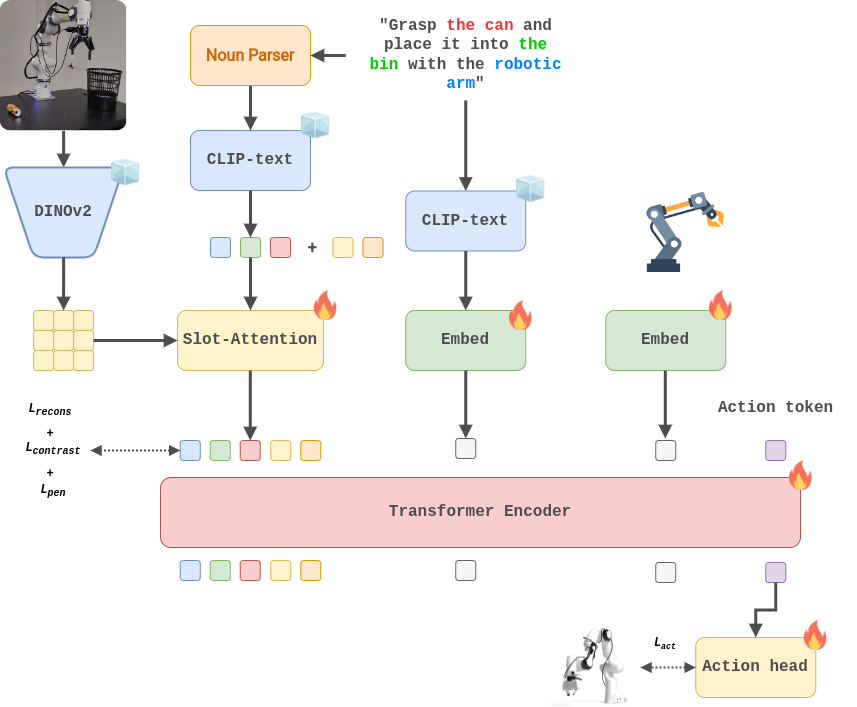}
        \caption{\textbf{Step 2: Dynamic task alignment}}
        \label{fig:method_robot_sub}
    \end{subfigure}
    
    \caption{\textbf{Overview of STORM.} STORM follows a two-stage training to produce task-aware object-centric representations for robotic control. 
    \textbf{(a) Semantic learning:} Frozen DINOv2 features are aggregated by a Slot-Attention module conditioned on noun embeddings extracted from text prompts using a frozen CLIP-text encoder. Reconstruction and contrastive losses encourage stable, semantically grounded slot formation while the penalization loss brings stability in the mask formation, avoiding degenerate solutions.  
    \textbf{(b) Dynamic task alignment:} The pretrained object-centric module extracts slots from camera observations, which are combined with task embeddings, robot proprioception, and a learnable \texttt{[ACT]} token and processed by a Transformer decoder policy. A GMM action head predicts the next action from the \texttt{[ACT]} token. The slot-attention module is further optimized with reduced learning rate to adapt the decomposition to the robotic dataset.}
    \label{fig:method_robot}
\end{figure*}

Our method, STORM (Fig.~\ref{fig:method_robot}), extends prior object-centric learning approaches~\cite{seitzer2023bridginggaprealworldobjectcentric,didolkar2025ctrlolanguagecontrollableobjectcentricvisual} by tailoring them specifically to the dynamic constraints of robotic control. Rather than simply applying existing models, we introduce three core methodological contributions to ensure robust and lightweight performance:
\begin{itemize}
    \item \textbf{A decoupled two-phase adaptation strategy:} We replace standard end-to-end training with a two-stage approach that stabilizes semantic slot discovery before further aligning them with policy gradients, preventing representation collapse.
    \item \textbf{Lightweight language conditioning \& Entropy regularization:} We eliminate the reliance on large language models for slot initialization, using only a lightweight CLIP-text encoder. To prevent the slot collapse that happens in this simplified setup, we introduce a novel entropy-based slot usage penalty. 
    \item \textbf{Task-aligned policy integration with spatial grounding:} We propose a policy architecture that explicitly combines task embeddings with slot representations. Crucially, we augment these slots with explicit spatial information derived from their masks (e.g., center of mass) to provide essential geometric cues and a dynamic alignment module permitting to have a stable slot formation through timesteps.
\end{itemize}

\subsection{Two-phase learning for object-centric adaptation}
While frozen VFM features provide strong visual representations, directly training object-centric layers and a control policy end-to-end frequently results in unstable or degenerate slot assignments. To overcome this, STORM's primary architectural contribution is a two-stage strategy that progressively introduces task supervision:

\begin{itemize}
    \item \textbf{Stage 1: Visual–Semantic Slot Pretraining (Section \ref{sec:overview}):} In the first stage, the object-centric module is trained entirely independently of the control policy. Given frozen DINOv2~\cite{oquab2024dinov2learningrobustvisual} feature maps, STORM produces a certain number of slots, each corresponding to a candidate object in the environment. We supervise these slots using a visual–semantic objective that aligns slot embeddings with CLIP text~\cite{radford2021learningtransferablevisualmodels} descriptions. An additional regularization loss encourages distinct, spatially localized assignments. Crucially, no policy gradients are used here, allowing the slots to freely map the visual-semantic landscape.
   \item \textbf{Stage 2: Joint Slot–Policy Training (Section \ref{sec:robot_meth}):} In the second stage, the pretrained object-centric module is integrated with a downstream policy via imitation learning. The object-centric module and the policy are optimized \emph{independently}: gradients from the policy loss are not backpropagated into the object-centric module, and a gradient detachment is applied at the slot level. However, the object-centric layers are fine-tuned to the domain of the policy with a small learning rate (using only images from the robotic dataset). This critical design choice allows the object-centric layers to preserve the structural and semantic biases acquired during Stage~1, while still adapting to task-relevant visual statistics.
\end{itemize}

\subsection{Visual-semantic slot pretraining}
\label{sec:overview}
The first training phase focuses on decomposing an input image into robust semantic entities based on textual prompts. As shown in the left part of Figure~\ref{fig:method_robot}, this framework comprises an object-centric decomposition branch and a language conditioning branch.

\paragraph{Object-centric decomposition}
Following established object-centric models~\cite{seitzer2023bridginggaprealworldobjectcentric}, an input image $I$ is encoded by a frozen DINOv2~\cite{oquab2024dinov2learningrobustvisual} backbone $B$ to extract patch features $F = \{f_1, .., f_N\} = B(I)$. A set of slots $S = \{s_1, ..., s_{K}\}$ is generally randomly initialized (in next-section we detail our slot initialization) and cross-attends to $F$ via Slot-Attention~\cite{locatello2020objectcentriclearningslotattention} (a modified cross-attention normalized over queries rather than keys). Finally, a shared MLP decoder $D$ reconstructs the DINOv2 feature maps $\hat{F}$ from the slots, optimized via : 
\begin{equation}
    \mathcal{L}_{recons} = ||F - \hat{F}||
\end{equation}
This loss indirectly guides the attention maps of the slots towards objects, thanks to the good property of the feature space of the input VFM. 

\paragraph{Lightweight language conditioning and semantic learning}
To control slot generation, prior works~\cite{didolkar2025ctrlolanguagecontrollableobjectcentricvisual} initialize slots using massive vision-language pipelines (e.g., combining a 7B-parameters LLM2Vec~\cite{behnamghader2024llm2veclargelanguagemodels} model with CLIP-text). To drastically reduce the memory and compute footprint for real-time robotic applications, \textbf{STORM entirely discards the LLM requirement}. Instead, we rely solely on a $\sim$300M parameter CLIP-text encoder~\cite{radford2021learningtransferablevisualmodels} (eg. a $\times23$ parameter reduction).

Rather than initializing slots randomly like in the original approach~\cite{locatello2020objectcentriclearningslotattention,seitzer2023bridginggaprealworldobjectcentric} they are initialized with the embedded names of the objects in the image with CLIP text. Then, each conditioned slot uses its associated soft mask $m_k$ (extracted from Slot-Attention layers) to pool image patches $F_{\text{CLIP}}$ from a dense CLIP visual encoder~\cite{fan2024unsupervisedopenvocabularyobjectlocalization}. A contrastive loss is applied between the mask-pooled visual features $F^{\text{masked}} = m_k \cdot F_{\text{CLIP}}$ and the CLIP text embeddings $z_{\text{emb}}$:
\begin{equation}
\mathcal{L}_{\text{sem}} = - \sum_{i=1}^M
\log \frac{\exp\left( z_i^{\text{emb}} \cdot F^{\text{masked}}_i / \tau \right)}
{\sum_{t=1}^{T} \exp\left( z_i^{\text{emb}} \cdot F^{\text{masked}}_t / \tau \right)},
\end{equation}
\label{eq:contrast}
where $M \leq K$ denotes the number of textual prompts.
We introduce an additional PatchCLIP branch, rather than directly aggregating DINOv2 features, because DINOv2 lacks an inherent vision-language alignment. PatchCLIP explicitly projects the localized visual patches into the same shared latent space as the CLIP text embeddings, enabling the direct computation of this semantic contrastive objective without requiring an additional learned projection.
During the first stage of semantic training, the prompt are obtained from labeled text-image pairs. 

\paragraph{Entropy regularization}
Because we removed the heavy LLM initialization, this contrastive loss alone can suffer from slot collapse (where attention collapses uniformly or onto a single slot). To mitigate this issue, we introduce a novel \textbf{entropy-based slot usage penalty}. Let $m_k \in \mathbb{R}^{B \times K \times N}$ denote the soft assignment masks (where $B$ is batch size, $K$ is slots, and $N$ is patches). We quantify slot collapse by aggregating usage, normalizing, and computing the entropy:
\begin{equation}
\begin{aligned}
    S_{b,k} &= \sum_{n=1}^{N} m_{b,k,n}, \\
    P_{b,k} &= \frac{S_{b,k}}{\sum_{j=1}^{K} S_{b,j} + \epsilon}, \\
    \mathcal{H}_b &= -\frac{1}{\log K} \sum_{k=1}^{K} P_{b,k} \log (P_{b,k} + \epsilon), \\
    \mathcal{L}_{\text{pen}} &= 1 - \frac{1}{B} \sum_{b=1}^{B} \mathcal{H}_b .
\end{aligned}
\end{equation}

Minimizing $\mathcal{L}_{\text{pen}}$ explicitly encourages balanced usage across all slots, serving as a critical regularizer for our lightweight formulation.

\textbf{Overall loss:}
The visual-semantic module is pre-trained by minimizing the combined objective:
\begin{equation}
    \mathcal{L}_{Overall} = \mathcal{L}_{recons} + 0.1 \times (\mathcal{L}_{sem} + \mathcal{L}_{pen})
\end{equation}
\label{eq:overall}

\subsection{Joint Slot-Policy Training}
\label{sec:robot_meth}
In the second phase, we integrate the semantically grounded object-centric representations into an imitation learning pipeline. Given expert demonstrations $\mathcal{D} = \{\tau_1, \ldots, \tau_n\}$ with trajectories $\tau_i = [(o_0,a_0), \ldots, (o_T,a_T)]$, a policy $\pi$ must predict actions $a_i$ from visual inputs $o_i$, task instructions, and proprioception. As illustrated in the right of Figure~\ref{fig:method_robot}, observations yield a fixed set of slot representations. Additionally, to further enhance spatial grounding of slots, a mask encoder extracts information from the masks associated to each slots to then concatenate this information to the slot embedding. This simple trick permits to further enhance the downstream results for robotic manipulation as discussed in~\ref{sec:abla}.

Task instructions are encoded via a frozen CLIP text encoder and provided as a global task embedding. To condition the object-centric module with nouns, rather than relying on manual dataset annotations or rigid rule-based heuristics, we automatically extract task-relevant entities by parsing nouns from the instruction using the pre-trained models in the spaCy natural language processing library~\cite{honnibalspacy}. Furthermore, we explicitly append ``robot arm'' to this list of parsed nouns to ensure the agent's embodiment is consistently distinguished from the rest of the scene elements. Proprioceptive inputs are projected as additional tokens. This entire sequence of tokens is processed by a Transformer decoder alongside a learnable \texttt{[ACT]} token, utilizing a 4-frame history to predict a chunk of 10 future actions. Finally, a Gaussian Mixture Model (GMM) action head maps the \texttt{[ACT]} token to the next relative joint command.

During joint training, we apply \textbf{feature-level gradient detachment} to isolate the learning signals. Specifically, we prevent the imitation learning loss from backpropagating into the visual backbone, meaning the policy is optimized exclusively via a standard imitation-learning objective. Concurrently, the object-centric components are independently \textbf{refined} using their own visual losses with a reduced learning rate.

\section{Experiments}
\label{sec:expe}
Following prior work on object-centric representation, we first validate our visual semantic learning by comparing it with existing object-centric models on classical \textbf{object-discovery} scenarios. Then, we evaluate our task-aware two-phase learning scheme on top of an existing frozen VFM to learn \textbf{robotic manipulation} tasks. We finally ablate the components of our framework to understand what constitutes the success of STORM. 

\subsection{Object decomposition and grounding}

\textbf{Pre-training details.} Our visual-semantic model described in Section~\ref{sec:overview} is pretrained for 300k steps on the VG-COCO dataset~\cite{didolkar2025ctrlolanguagecontrollableobjectcentricvisual} using the AdamW optimizer~\cite{loshchilov2019decoupledweightdecayregularization}. We use a learning rate of $4\times10^{-4}$ with a cosine decay schedule, 10{,}000 warmup steps, a batch size of 64 on a single V100 GPU. We set the number of slots to 7 with a slot dimension of 256. We use the DINOv2-B/14 version.

\textbf{Metrics.} Object discovery performance is commonly evaluated using the \textit{Foreground Adjusted Rand Index} (FG-ARI)~\cite{Hubert1985}, which quantifies segmentation accuracy and object-consistency, and \textit{Mean Best Overlap} (mBO)~\cite{Pont_Tuset_2017}, which measures the pixel-wise overlap between predicted and ground-truth object masks. 

\textbf{Object discovery setup.} We benchmark our visual-semantic model on object-centric discovery tasks using PASCAL VOC 2012~\cite{Everingham2010} and COCO~\cite{lin2015microsoftcococommonobjects} datasets, both of which contain complex, multi-object scenes. These datasets are standard in the literature, allowing direct comparison with existing object-centric methods. It is important to emphasize that this evaluation serves primarily as a validation step for our visual-semantic module. Rather than aiming for state-of-the-art 2D segmentation, our goal here is to confirm that our lightweight, LLM-free architecture effectively separates distinct entities without degrading the semantic grounding capabilities seen in heavier, more complex models. We compare STORM against several baselines: DINOSAUR \cite{seitzer2023bridginggaprealworldobjectcentric}, Stable-LSD~\cite{jiang2023objectcentricslotdiffusion}, Slot-Diffusion~\cite{wu2023slotdiffusionobjectcentricgenerativemodeling}, and our closest baseline, CTRL-O~\cite{didolkar2025ctrlolanguagecontrollableobjectcentricvisual}. CTRL-O is also weakly-supervised but relies on a much stronger, compute-heavy semantic model as input (e.g., LLM2Vec) to initialize its slots.

For all models, masks are extracted from the attention maps within the Slot-Attention module and resized to the original image dimensions before being evaluated against the ground-truth masks.

\paragraph{Results}
For all baselines, we report the results as provided in their respective papers (Table~\ref{table:perf_object_discovery}). STORM surpasses all unsupervised models in terms of FG-ARI and it performs highly competitively with CTRL-O. This successfully validates our design choices: despite dropping the massive language models used by CTRL-O, our module maintains strong object discovery performances. While our model slightly lags in mBO compared to some diffusion-based methods, it clearly exceeds them in segmentation quality (FG-ARI). Note that STORM deliberately uses the original, lightweight MLP decoder from DINOSAUR, which is known to produce less precise masks than heavy transformer or diffusion decoders. Integrating such decoders could further enhance mask sharpness, but our results confirm that the learned slots are semantically accurate and structurally sound for initiating training on downstream robotic control.

\begin{table}[ht]
\centering
\caption{\textbf{Object discovery performance.} Metrics include mBO$^i$ (instance-level), mBO$^c$ (class-level), and FG-ARI. Bold is the best result, underline is the second best. Unsupervised (U), Weakly-Supervised (WS).}
\label{table:perf_object_discovery}

\vspace{0.5em}
\textbf{(a) Pascal VOC} \\
\vspace{0.2em}
\begin{tabular}{c | l  c c c}
\hline
\textbf{Sup.} & \textbf{Model} & \textbf{mBO$^i$} & \textbf{mBO$^c$} & \textbf{FG-ARI} \\
\hline
\multirow{4}{*}{U}
 & DINOSAUR \cite{seitzer2023bridginggaprealworldobjectcentric} & 39.5 & 40.9 & \underline{24.6} \\
 & Stable-LSD \cite{jiang2023objectcentricslotdiffusion} & -- & -- & -- \\
 & Slot-Diffusion \cite{wu2023slotdiffusionobjectcentricgenerativemodeling} & \textbf{50.4} & \textbf{55.3} & 17.8 \\
\hline
\multirow{2}{*}{WS}
 & CTRL-O \cite{didolkar2025ctrlolanguagecontrollableobjectcentricvisual} & -- & -- & -- \\
 & \cellcolor{gray!30} \textbf{STORM (Ours)} & \cellcolor{gray!30} \underline{40.6} & \cellcolor{gray!30} \underline{46.5} & \cellcolor{gray!30} \textbf{45.8} \\
\hline
\end{tabular}

\vspace{1.5em} 

\textbf{(b) COCO} \\
\vspace{0.2em}
\begin{tabular}{c | l c c c}
\hline
\textbf{Sup.} & \textbf{Model} & \textbf{mBO$^i$} & \textbf{mBO$^c$} & \textbf{FG-ARI} \\
\hline
\multirow{4}{*}{U}
 & DINOSAUR \cite{seitzer2023bridginggaprealworldobjectcentric} & 27.7 & 30.9 & 40.3 \\
 & Stable-LSD \cite{jiang2023objectcentricslotdiffusion} & \underline{30.4} & -- & 35.0 \\
 & Slot-Diffusion \cite{wu2023slotdiffusionobjectcentricgenerativemodeling} & \textbf{31.0} & \underline{35.0} & 37.2 \\
\hline
\multirow{2}{*}{WS}
 & CTRL-O \cite{didolkar2025ctrlolanguagecontrollableobjectcentricvisual} & 27.2 & -- & \textbf{47.5} \\
 & \cellcolor{gray!30} \textbf{STORM (Ours)} & \cellcolor{gray!30} 27.8 & \cellcolor{gray!30} \textbf{36.6} & \cellcolor{gray!30} \underline{44.1} \\
\hline
\end{tabular}
\end{table}

\subsection{Robotic manipulation}
\label{sec:robot}

\paragraph{Baselines.} To evaluate our approach, we compare against several visual baselines. \textbf{DINOv2~\cite{oquab2024dinov2learningrobustvisual}:} A state-of-the-art VFM providing dense, general-purpose patch features. We evaluate both frozen and fine-tuned variants. \textbf{DINOSAUR~\cite{seitzer2023bridginggaprealworldobjectcentric}:} A leading unsupervised object-centric model that uses Slot Attention to reconstruct pre-trained vision transformer features with unguided slot formation (without any task-awareness). \textbf{SAM + DINOv2~\cite{shi2024composingpretrainedobjectcentricrepresentations}:} An explicit "segment-then-extract" approach that extracts 2D object masks using SAM and pools DINOv2 features for each region.
DINOSAUR serves as our purely visual, task-agnostic baseline; comparing STORM against it highlights the necessity of our semantic language conditioning and two-phase training strategy to prevent slots from latching onto task-irrelevant background elements. Comparing against DINOv2 isolates the benefit of object-centric bottleneck, demonstrating how discrete representations improve robustness to visual distractors over dense features. Finally, the SAM + DINOv2 baseline allows us to contrast explicit hard-segmentation methods, against STORM's continuous, differentiable soft-attention slots.

In our robotic manipulation experiments, all the pre-trained object-centric layers are fine-tuned in parallel with the policy (Slot-Attention, language conditionning and slot decoder), but the backbone (DINOv2) is kept frozen (unless specified otherwise). Joint-training is performed for 150k steps directly on the example trajectories. The visual loss is identical to that used during pre-training, but with a reduced learning rate of $1e^{-5}$ all other hyperparameters for the visual module remain unchanged. The policy is optimized using an MSE loss between the normalized predicted actions and the ground-truth actions. 

Importantly, the same policy architecture is shared across all visual models and follows the structure detailed in Section~\ref{sec:method}. All training hyperparameter for policy training are shared amongst different baselines. 

\begin{figure}
    \centering
    \includegraphics[width=0.85\linewidth]{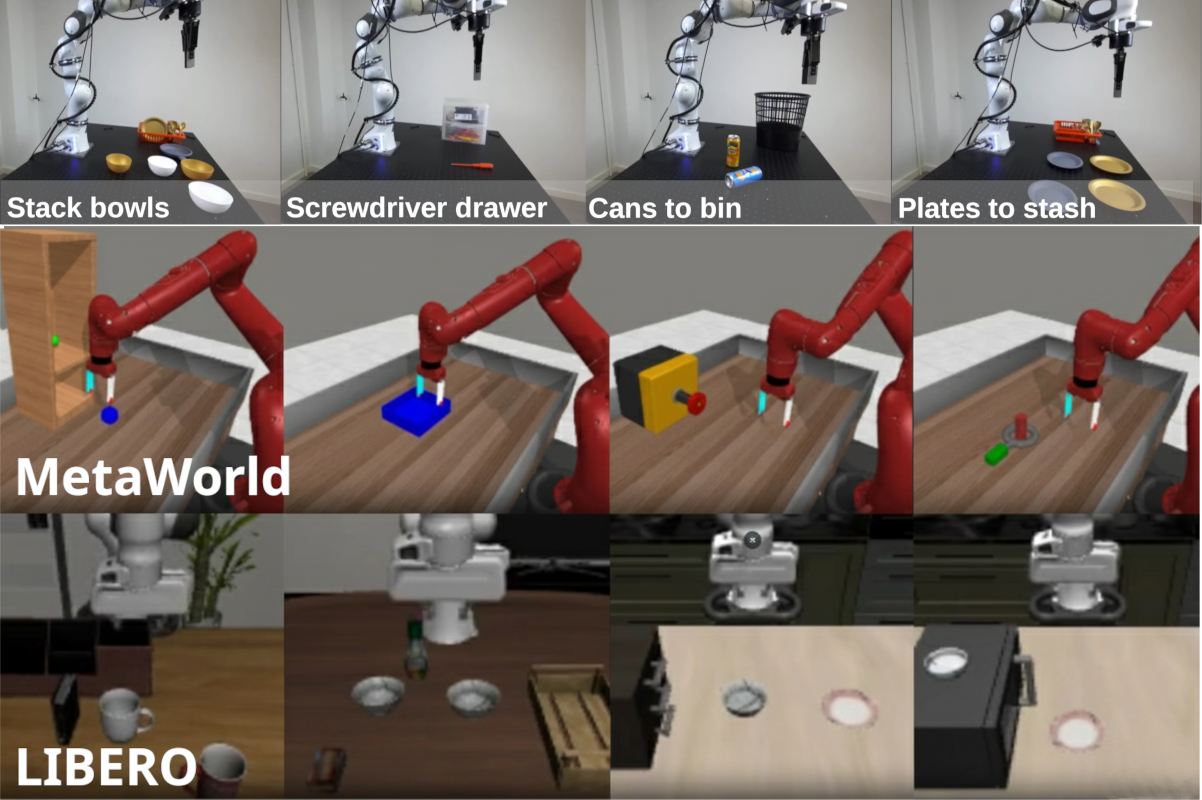}
    \caption{\textbf{Robotics environments visualization.} Examples of the environments used in our experiments: Real-world tasks (top row), MetaWorld (middle row) and LIBERO (bottom row).}
    \label{fig:environments}
\end{figure}

\paragraph{Evaluation Benchmarks.} To comprehensively assess our approach, we evaluate the models across three diverse environments depicted on Figure~\ref{fig:environments}, testing both in-distribution (ID) performance and out-of-distribution (OOD) generalization:
\begin{itemize}
    \item \textbf{MetaWorld~\cite{mclean2025metaworldimprovedstandardizedrl}:} A simulated robotic manipulation environment with a Sawyer arm. We evaluate performance across a subset of 10 tasks following~\cite{chapin2025objectcentricrepresentationsimprovepolicy} setup.
    \item \textbf{LIBERO~\cite{liu2023libero}:} A simulated suite of a Franka robot arm that introduces higher visual complexity and task diversity compared to MetaWorld. It is composed of 90 diverse tasks implying multi-object scenes and diverse environments such as an office or a kitchen.
    \item \textbf{Real-world scenarios:} We deploy the policies on a Franka robot to validate our simulation findings under real physical constraints. This ensures the learned object-centric representations effectively transfer to the physical world and maintain control reliability even under challenging, real-world visual distribution shifts. We use 4 diverse manipulation tasks: stacking bows into a pan, opening a drawer putting a screwdriver inside and closing the drawer, putting cans to a bin and placing plates into a dish rack.  
\end{itemize}

\begin{figure}
    \centering
    \includegraphics[width=0.85\linewidth]{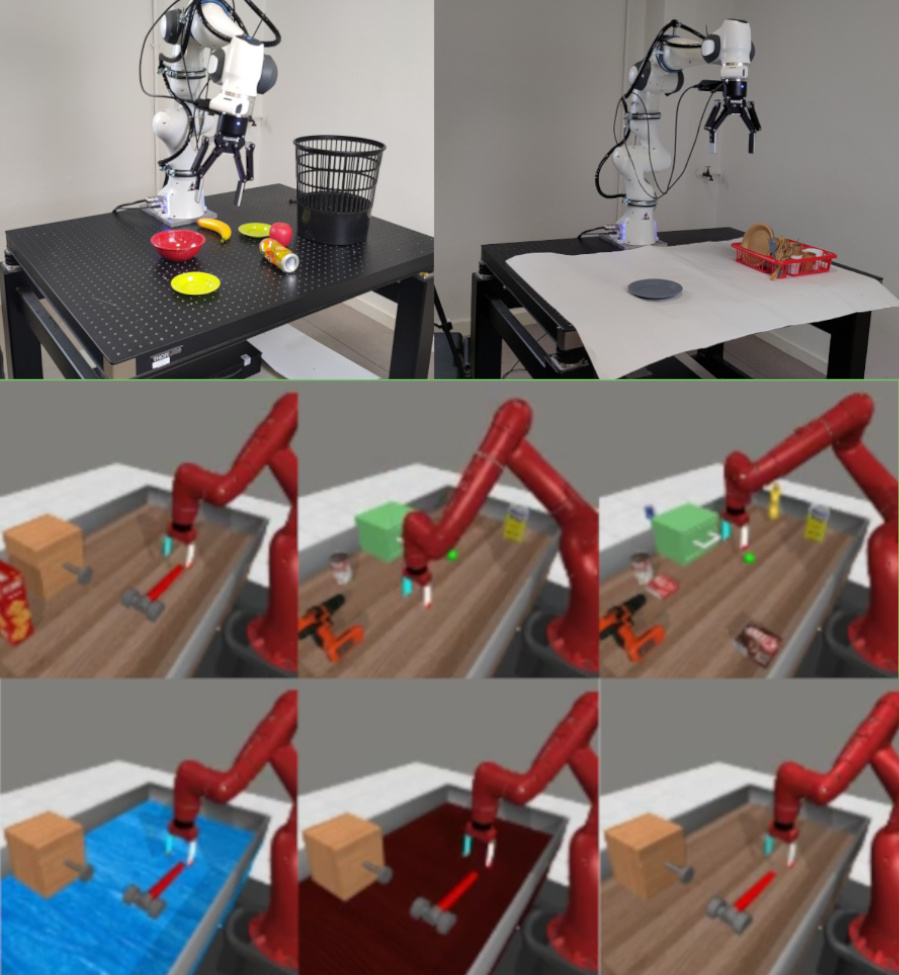}
    \caption{\textbf{Generalization visualization MetaWorld \& Real-world.} Examples of the generalization levels for real-world (new distractors, new texture) and MetaWorld (new distractors, new lighting, new texture).}
    \label{fig:generalization}
\end{figure}

For all experiments, 50 demonstrations are collected per task (in teleoperation for real-world). In simulation, we evaluate and apply a mean over 50 rollouts per task while we perform the evaluation on 15 rollouts per tasks on the real-world robot. Performance is evaluated using binary sucess-rate per rollout. The tasks is considered as a success if the object to be manipulated is placed within the goal area it is supposed to be left. In the real world experiments it is a success if all bowls are stacked in the pan, the scredriver is inside the closed drawer, the cans are inside the bin and the plates are all correctly placed in the dishrack. Figure~\ref{fig:generalization} showcases the various generalization levels across both real-world and simulated environments.

\paragraph{Results}
Table~\ref{tab:gen_policies} reports the overall manipulation performance on MetaWorld~\cite{mclean2025metaworldimprovedstandardizedrl}, LIBERO~\cite{liu2023libero}, and real-world scenarios, evaluated both in-distribution (ID) and under out-of-distribution (OOD) conditions (novel distractors, novel lighting conditions, novel textures). 

On MetaWorld, we observe a clear hierarchy between visual representations. The frozen DINOv2 baseline performs correctly in ID settings (73.8\%) but suffers a severe degradation under visual shifts, dropping 34.2 points to 39.6\% in the OOD setting. Fine-tuning the backbone (DINOv2 ft.) slightly improves robustness (+4.2 points in OOD) but at the cost of ID performance, suggesting a trade-off between specialization and generalization. DINOSAUR performs similarly to the frozen baseline in ID (73.2\%) while offering improved OOD generalization (46.0\%). This already suggests that object-centric layers are an interesting way of providing robustness to the original frozen VFM. Conversely, explicitly combining SAM and DINOv2 results in poor performance (45.8\% ID, 15.2\% OOD), likely due to the lack of explicit tuning of local features. In contrast, STORM achieves the best performance across both settings. It not only maintains strong ID accuracy (74.8\%) but also substantially boosts robustness to unseen visual conditions, improving the OOD success rate by 12.7 points over the frozen baseline.

On LIBERO, the performance gap becomes even more pronounced. While the frozen DINOv2 baseline maintains a respectable success rate of 78.9\% (ID) and 70.3\% (OOD), fine-tuning the backbone (DINOv2 ft.) actually results in slight performance drops in both settings. This suggests that standard fine-tuning may struggle with the higher visual complexity and task diversity present in the LIBERO suite. DINOSAUR remains competitive but fails to surpass the frozen baseline. SAM+DINOv2 showcase again a huge drop in performances. STORM, however, outperforms the best baseline by 10.7 points in ID settings and 19.0 points under OOD conditions. Notably, STORM exhibits almost no performance decay when transitioning from ID to OOD environments on LIBERO (89.6\% vs. 83.1\%), whereas the baselines show a more noticeable gap. This suggests that bottleneck in STORM allow the policy to remain focused on task-relevant features, effectively ignoring visual perturbations that typically degrade other representations.

Finally, the real-robot evaluations further validate these findings under physical constraints. The frozen DINOv2 baseline struggles significantly, achieving 22.0\% ID and near-zero success (0.1\%) in OOD. While fine-tuning (DINOv2 ft.) and DINOSAUR offer noticeable improvements,  reaching 40.2\% and 40.1\% in ID, STORM demonstrates a substantial boost. It achieves a 55.2\% success rate in ID (+33.2 points over frozen DINOv2) and a highly robust 41.9\% in OOD (+41.8 points). This highlights that the object-centric representations learned by STORM are highly effective for transferring capabilities to the physical world, even under challenging visual distribution shifts.

\begin{table*}[h]
\centering
\caption{\textbf{Overall evaluations} Success rate (\%) on MetaWorld, LIBERO and real-world benchmarks. We report performance both In-Distribution (ID) and under Out-Of-Distribution (OOD) conditions. Bold is the best result, underline is the second best. Relative performance to DINOv2 frozen is shown in (green) and (red).}
\label{tab:gen_policies}
\resizebox{0.7\linewidth}{!}{%
\setlength{\tabcolsep}{4pt}
\begin{tabular}{l| c c | c c| c c}
\hline
\multirow{2}{*}{\textbf{Visual model}} 
& \multicolumn{2}{c|}{\textbf{MetaWorld}} 
& \multicolumn{2}{c|}{\textbf{LIBERO}}
& \multicolumn{2}{c}{\textbf{Real-world}} \\
\cline{2-7}
& ID~$\uparrow$ & OOD~$\uparrow$ & ID~$\uparrow$ & OOD~$\uparrow$ & ID~$\uparrow$ & OOD~$\uparrow$ \\
\hline
DINOv2\ding{100}& \underline{73.8} & 39.6 & \underline{78.9} & \underline{70.3} & 22.0 & 0.1 \\

DINOv2 (ft.) & 71.8 {\scriptsize \gradcolor{-2.0}(-2.0)} & \underline{43.8} {\scriptsize \gradcolor{4.2}(+4.2)} & 77.5 {\scriptsize \gradcolor{-1.4}(-1.4)} & 65.3 {\scriptsize \gradcolor{-5.0}(-5.0)} & \underline{40.2} {\scriptsize \gradcolor{18.2}(+18.2)} & \underline{12.4} {\scriptsize \gradcolor{12.3}(+12.3)} \\

\hline

SAM + DINOv2 & 45.8 {\scriptsize \gradcolor{-28.0}(-28.0)} & 15.2 {\scriptsize \gradcolor{-24.4}(-24.4)} & 31.0 {\scriptsize \gradcolor{-47.9}(-47.9)} & 20.5 {\scriptsize \gradcolor{-49.8}(-49.8)} & 0.0 {\scriptsize \gradcolor{-22.0}(-22.0)} & 0.0 {\scriptsize \gradcolor{-0.1}(-0.1)} \\

DINOSAUR & 73.2 {\scriptsize \gradcolor{-0.6}(-0.6)} & 46.0 {\scriptsize \gradcolor{6.4}(+6.4)} & 77.3 {\scriptsize \gradcolor{-1.6}(-1.6)} & 73.2 {\scriptsize \gradcolor{2.9}(+2.9)} & 40.1 {\scriptsize \gradcolor{18.1}(+18.1)} & 27.2 {\scriptsize \gradcolor{27.1}(+27.1)} \\
\hline
\rowcolor{gray!30} \textbf{STORM (Ours)} & \textbf{74.8} {\scriptsize \gradcolor{1.0}\textbf{(+1.0)}} & \textbf{52.3} {\scriptsize \gradcolor{12.7}\textbf{(+12.7)}} & \textbf{89.6} {\scriptsize \gradcolor{10.7}\textbf{(+10.7)}} & \textbf{83.1} {\scriptsize \gradcolor{19.0}\textbf{(+13.2)}} & \textbf{55.2} {\scriptsize \gradcolor{33.2}\textbf{(+33.2)}} & \textbf{41.9} {\scriptsize \gradcolor{41.8}\textbf{(+41.8)}} \\
\hline
\end{tabular}%
}
\end{table*}

\subsection{Ablation studies}
\label{sec:abla}
We conduct ablations to evaluate the effects of STORM's key components. Specifically, we investigate: (1) the impact of the two-phase learning, (2) the impact of the novel task-awareness module, (3) the influence of different masks representations. All of our ablative studies are performed on the MetaWorld~\cite{mclean2025metaworldimprovedstandardizedrl} benchmark. Results detailed in Table~\ref{table:ablation} are analyzed in following paragraphs.

\paragraph{Two-phase learning \& Task awareness} Naïvely training task-aware slots from scratch directly on the downstream manipulation data leads to a significant performance drop in both ID and OD settings. This degradation indicates that jointly learning object decomposition and policy control from limited task-specific demonstrations is unstable and detrimental. 

Pre-training the task-aware slot-based object-centric layers and keeping them frozen during the policy training partially alleviates this issue, yielding comparable ID performance to the frozen VFM and a notable improvement in OD generalization. 

Removing the task-aware training aligning slots with task-embeddings leads back to the DINOSAUR setup. This model decompose scenes into discrete entities but lack an inherent understanding of which objects are relevant to the current objective. DINOSAUR achieves a respectable 73.2\% in ID, but its performance drops to 46.0\% under OD conditions.

Nevertheless, the best results are achieved with our proposed task-aware two-phase training strategy (STORM), where object-centric representations are first learned independently and then carefully adapted during policy training. This decoupling stabilizes optimization and enables the slots to maintain coherent object bindings while adapting to task-specific cues, resulting in consistent improvements in both ID and OD performance.

\begin{table}[h]
\caption{\textbf{Effect of object-centric layers and multi-phase learning.} 
Success Rate (SR) (\%) on \textbf{MetaWorld}. \textbf{ID} corresponds to the standard benchmark, 
while \textbf{OD} corresponds to out-of-distribution evaluations. 
Relative performance to DINOv2 is shown with color gradients.}
\label{table:ablation}
\begin{center}
\setlength{\tabcolsep}{4pt}
\begin{tabular}{cccc}
\hline
\multirow{2}{*}{\textbf{Model}} & \multirow{2}{*}{\textbf{Pre-train}} & 
\multicolumn{2}{c}{\textbf{Success Rate (\%)}} \\ 
\cline{3-4}
 & & \textbf{ID} & \textbf{OD} \\
\hline
DINOv2\ding{100}~\cite{oquab2024dinov2learningrobustvisual}  
& -- & 73.8 & 39.6 \\

DINOv2\ding{100} + Slots scratch 
& \xmark 
& 69.0 {\scriptsize \gradcolor{-4.8}(-4.8)} 
& 32.5 {\scriptsize \gradcolor{-7.1}(-7.1)} \\

DINOSAUR 
& \cmark 
& 73.2 {\scriptsize \gradcolor{-1.5}(-1.5)} 
& 46.0 {\scriptsize \gradcolor{6.4}(+6.4)} \\

DINOv2\ding{100} + Slots\ding{100} 
& \cmark 
& \underline{74.4} {\scriptsize \gradcolor{+0.6}(+0.6)} 
& \underline{48.4} {\scriptsize \gradcolor{8.8}(+8.8)} \\

\hline
\rowcolor{gray!10}
\textbf{STORM (Ours)} 
& \cmark 
& \textbf{74.8} {\scriptsize \gradcolor{1.0}\textbf{(+1.0)}} 
& \textbf{52.3} {\scriptsize \gradcolor{12.7}\textbf{(+12.7)}} \\
\hline
\end{tabular}
\end{center}
\end{table}

\begin{table}[h]
\caption{\textbf{Effect of mask representation.} Success Rate (SR) (\%) on \textbf{MetaWorld}. \textbf{ID} corresponds to the standard benchmark, while \textbf{OD} corresponds to out-of-distribution evaluations. Bold is the best result, underline is the second best. Relative performance to $\emptyset$ is shown with color gradients.}
\label{tab:abla_mask}
\begin{center}
\setlength{\tabcolsep}{4pt}
\begin{tabular}{ccc}
\hline
\multirow{2}{*}{\textbf{Mask repre.}} & \multicolumn{2}{c}{\textbf{MetaWorld SR (\%)}} \\
\cline{2-3}
& \textbf{ID} & \textbf{OD} \\
\hline
$\emptyset$ 
& 64.4 & 45.5 \\

\texttt{bbox} 
& 68.8 {\scriptsize \gradcolor{4.4}(+4.4)} 
& 46.2 {\scriptsize \gradcolor{0.7}(+0.7)} \\

\texttt{mask} 
& \underline{69.4} {\scriptsize \gradcolor{5.0}(+5.0)} 
& \textbf{52.8} {\scriptsize \gradcolor{7.3}\textbf{(+7.3)}} \\

\hline
\rowcolor{gray!10}
\texttt{center} 
& \textbf{74.8} {\scriptsize \gradcolor{10.4}\textbf{(+10.4)}} 
& \underline{52.3} {\scriptsize \gradcolor{6.8}(+6.8)} \\
\hline
\end{tabular}
\end{center}
\end{table}

\paragraph{Mask representation:} In addition to learning object-centric slots, we provide the policy with explicit spatial information derived from the slot masks. Mask representations encode complementary geometric cues (object's location, extent, and shape) that are not fully captured by appearance features alone. We evaluate three different mask encodings: \texttt{center}, which uses the mask’s center of mass; \texttt{bbox}, which encodes the bounding box coordinates; and \texttt{mask}, which directly embeds the binary mask using a shallow CNN.

Table~\ref{tab:abla_mask} shows that incorporating mask information is crucial for strong performance. Removing mask cues leads to an important drop in success rates, confirming that spatial grounding is essential for manipulation tasks. Among the evaluated representations, the simple \texttt{center} encoding performs best on MetaWorld, outperforming more expressive alternatives, and is comparable to \texttt{mask} on MetaWorld-OOD. This suggests that a compact and stable representation of object position is sufficient, and even preferable, for policy learning. While directly encoding the full mask also yields strong results, it introduces additional complexity that does not consistently translate into higher performance. In contrast, bounding box representations appear too coarse and sensitive to noise, offering limited benefit over not using mask information at all. In all our ablations and evaluations, all object-centric baselines used the \texttt{center} encoding as masks spatial information for a fair comparison.

\section{Conclusion}
We presented STORM, a compact, slot-based module that adapts frozen foundation model features into task-aware, object-centric representations for robotic manipulation. The key insight is that a \emph{two-phase} training strategy: semantic slot pretraining followed by cautious joint adaptation with a policy, stabilizes slot formation and yields representations that are both semantically meaningful and useful for control. Empirically, STORM consistently improves robustness to visual distractors and enhances generalization compared to frozen and naively fine-tuned baselines on simulated and real-world manipulation benchmarks. Our ablations show that minimal spatial encodings (e.g., slot centers) are an effective and robust interface between perception and action, and that staged learning is essential to avoid degenerate slot assignments when adapting object-centric layers for control. Looking forward, extending STORM to variable slot counts, multi-scale part decomposition will be important steps for deploying object-centric, language-aware perception in real robotic systems. 

We hope this work provides a practical path for leveraging more efficiently large VFM in control: rather than tuning large backbones entirely, small and semantically-grounded adapters trained with multi-phase strategies can yield strong and robust downstream behavior.

\section*{ACKNOWLEDGMENT}
This work was in part supported by the French Research Agency, l'Agence Nationale de Recherche (ANR), through the projects Chiron (ANR-20-IADJ-0001-01), Aristotle (ANR-21-FAI1-0009-01), Astérix (ANR-23-EDIA-0002), Demeter (ANR-25-HTCE-0002) and Protheus ( ANR-25-TSIA-0011-01), the French national investment prioritary program PSPC FAIR WASTE project, the French-Singaporean collaborative Embodied AI project within the CREATE, as well as a donation to Fonds de Dotation Centrale Lyon by Huawei Technologies R\&D France. It was granted access to the HPC resources of IDRIS under the allocation 2025-[AD011016842], 2025-[AD011016615] and 2026-[A0201017513] made by GENCI.

\end{document}